\documentclass[10pt,twocolumn,letterpaper]{article}

\usepackage{wacv}
\usepackage{times}
\usepackage{epsfig}
\usepackage{graphicx}
\usepackage{amsmath}
\usepackage{amssymb}
\usepackage[para,online,flushleft]{threeparttable}

\usepackage{pdfpages}
\usepackage{subcaption}
\usepackage[table,xcdraw]{xcolor}
\usepackage{tabularx} 
\usepackage{commath}
\usepackage{placeins}
\newcommand{\vect}[1]{{\bf {#1}}}
\newcommand{\matx}[1]{{\cal {#1}}}
\newcommand{\expect}[1]{{\mathbb{E}}{{\left[{{#1}}\right]}}}
\DeclareMathOperator*{\argmin}{argmin}

\newcommand{\ea}[0]{{\em et al. }}

\wacvfinalcopy 

\def\wacvPaperID{****} 

\ifwacvfinal\fi
\begin{document}

\title{Improving Style Transfer with Calibrated Metrics}

\author{Mao-Chuang Yeh\thanks{ First two authors have equal contribution}~~~~~~~~~Shuai Tang\footnotemark[1]~~~~~~~~~Anand Bhattad~~~~~~~~~Chuhang Zou~~~~~~~~~David Forsyth\\
University of Illinois at Urbana-Champaign\\
\{myeh2, stang30, bhattad2, czou4, daf\}@illinois.edu}

\maketitle
\ifwacvfinal\thispagestyle{empty}\fi

\begin{abstract}

Style transfer produces a transferred image which is a rendering of a content image in the manner of a style image. We seek to understand how to improve style transfer.

To do so requires quantitative evaluation procedures, but current evaluation is qualitative, mostly involving user studies. We describe a novel quantitative evaluation procedure.   Our procedure relies on two statistics: the Effectiveness (E) statistic measures the extent that a given style has been transferred to the target, and the Coherence (C) statistic measures the extent to which the original image's content is preserved.  Our statistics are calibrated to human preference: targets with larger values of E and C will reliably be preferred by human subjects in comparisons of style and content, respectively.

We use these statistics to investigate relative performance of a number of Neural Style Transfer (NST) methods, revealing a number of intriguing properties.  Admissible methods lie on a Pareto frontier (i.e. improving E reduces C, or vice versa). Three methods are admissible: Universal style transfer produces very good C but weak E; modifying the optimization used for Gatys' loss produces a method with strong E and strong C; and a modified cross-layer method has slightly better E at strong cost in C.  While the histogram loss improves the E statistics of Gatys' method, it does not make the method admissible.  Surprisingly, style weights have relatively little effect in improving EC scores, and most variability in transfer is explained by the style itself (meaning experimenters can be misguided by selecting styles). Our GitHub Link is available. \footnote{ https://github.com/stringtron/quantative\_style}
\end{abstract}


\begin{figure}
    \centering
    \includegraphics[width=0.9\linewidth]{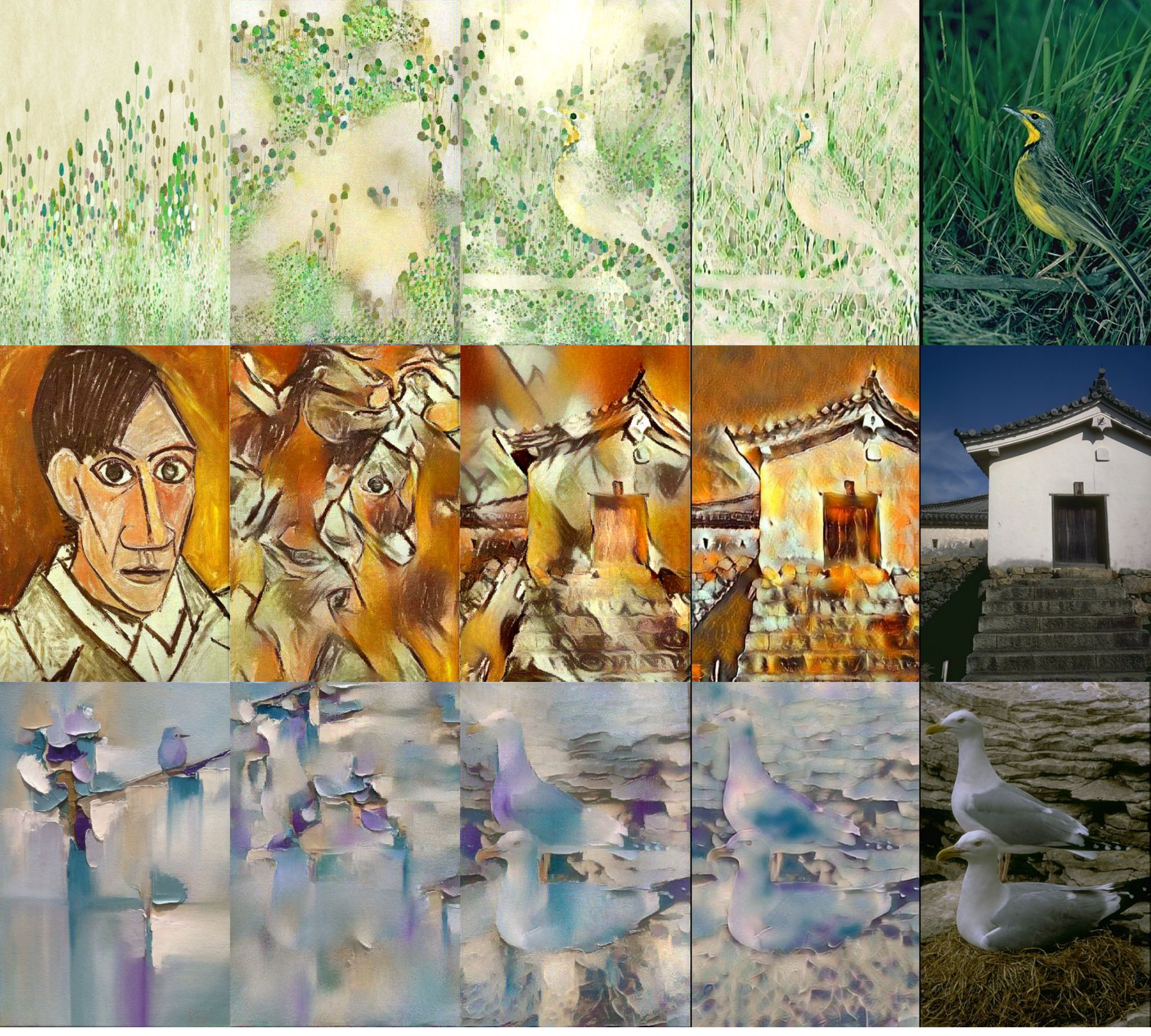}
\caption{A grid of stylized images visualizing the Effectiveness-Coherence space. From left to right, each row shows \textit{style image}, \textit{XLCM}, \textit{GAL}, \textit{Universial} and \textit{content image} ~(see method details in Sec.\ref{ExpPro}) qualitative results for the same style-content pair. Note for the three example transfer methods, from left to right, the Effectiveness scores decrease and the Coherence scores increase. Also note all images are sampled near their method's EC mean which is on "Pareto-optimal curve" of all compared transfer methods. }
\label{visspace}
\end{figure}

\vspace{-0.6cm}
\section{Introduction}


In this paper, 
we seek to identify factors that lead to better style transfers. To do so, we construct a comprehensive quantitative evaluation procedure for style transfer methods. We evaluate style transfers on two criteria.  {\bf Effectiveness} (E) measures whether transferred images have the desired style, using divergence between Convolutional Neural Network (CNN) feature layer distributions of the synthesized image and original image.
{\bf Coherence} (C) measures whether the synthesized images respect the underlying decomposition of the content image into
objects, using established contour detection procedures together with the colored natural images from BSDS500 dataset 
\cite{arbelaez2011contour}. Both our E and C measures are calibrated by user studies in Sec. 4.

 Our qualitative metric mainly focuses on the analysis of Parametric Neural Methods (under the taxonomy of NST techniques)~\cite{jing2019neural}. 
The non-Parametric Methods may generate a largely different feature statistics from original style image due to the pattern fitting to the content image, which are intrinsically different from Parametric ones. Therefore, it is not necessary to evaluate two types of transfer methods by the same quantitative metric at this stage.


{\bf Contributions:} We present E and C measures of style transferred images (see Fig.~\ref{visspace}). Our
measures are highly effective at predicting user preferences. We use our measures to compare several style transfer
methods quantitatively.  Our study suggests that controlling
cross-layer loss is helpful, particularly if one uses the cross-layer
covariance matrix (rather than Gram matrix).  Our study suggests that, despite the analysis of Risser \ea~\cite{risser2017stable},
the main problem with Gatys' method is optimization rather than symmetry; modifying the optimization leads to an
extremely strong method.  Gatys' method is unstable with high style weights, and we construct explicit models of the symmetry groups for Gatys' style loss and the cross-layer style loss 
(improving over Risser \ea, who could not construct the groups), which may explain this effect.   Our study suggests that, even for the best methods we
investigated, the effect of choice of style image is strong, meaning that it is dangerous for experimenters to select
style images when reporting results.
\vspace{-0.3cm}
\section{Related work} \label{sec:gatys}
\textbf{Style transfer}: bilinear models~\cite{Tenenbaum2000} , non-parametric methods ~\cite{Efros2001}, image analogies~\cite{Hertzmann2001} and adjusting filter statistics ~\cite{debonet,simoncelli} are capable of image style transfer and yield texture synthesis.  Gatys \ea demonstrated that producing neural network layers with particular
summary statistics (i.e. Gram matrices) yielded effective texture synthesis~\cite{NIPS2015_5633}. 
Gatys \ea achieved style transfer by searching for an image that satisfies both style texture summary statistics and
content constraints~\cite{gatys2016image}. This work has been much elaborated ~\cite{Johnson2016Perceptual, wang2016multimodal,chen2017stylebank,dumoulin2016learned, DBLP:journals/corr/UlyanovVL16,huang2017arbitrary,UST,li2018closed,chen2016fast,Shih2014,Luan2017,gatys2016controlling,li2017demystifying,champandard2016semantic,jing2017neural}. Novak and Nikulin noticed that cross-layer  
 Gram matrices reliably produce improvement on style transfer (\cite{novak2016improving}). However, 
 their work was an exploration of variants of style transfer rather than a thorough study to gain insights on 
 style summary statistics; since then, the method has been ignored in the literature.

\textbf{Style transfer evaluation}: style transfer methods are currently evaluated mostly by visual inspection on a small set of different styles and content image pairs. To our knowledge, there are no quantitative protocols to
evaluate the competence of style transfer apart from user studies ~\cite{li2018closed} (who also investigate edge coherence between content and stylized images).

\textbf{Gram matrices symmetry} in a style transfer loss function occur when there is a transformation available that changes the style transferred image without changing the value of the loss function. Risser \ea note instability in Gatys' method; symptoms are: poor and good style transfers of the same style to the same content with about the same loss value~\cite{risser2017stable}. They supply evidence that this behavior can be controlled by adding a histogram loss, which breaks the symmetry.  They do not write out the symmetry group as too complicated (~\cite{risser2017stable}, p 4-6).  Gupta \ea ~\cite{gupta2017characterizing} link instability in Gaty's method to the size of the trace of the Gram matrix.

\subsection{Gatys Method and Notation}
We review the original work of Gatys \ea \cite{gatys2016image} in detail to introduce notation.
Gatys finds an image where early layers of convolutional features match the lower layers of the style image and higher layers match the higher layers of a content image.  Write $I_{s}$ for the style, $I_{c}$, $I_{n}$ for the content and the new image, respectively,
and $\alpha$ for some parameters balancing style and content losses ($L_s$ and $L_c$ respectively).  Occasionally, we
will write $I_n^m(I_c, I_s)$ for the image resulting from style transfer using method $m$ applied to the arguments.
We obtain $I_{n}$ by finding
\begin{equation}
   {\argmin_{I_n}} L_c(I_{n}, I_{c})+\alpha L_s(I_{n}, I_{s})
\end{equation}
Losses are computed on a network representation, with $L$ convolutional layers, where the $l$'th layer
produces a feature map $f^l$ of size $H^l \times W^l \times C^l$ for height, width, and channel number, respectively.  We partition
the layers into three groups (style, content and target). Then we reindex the spatial variables (height and width) and
write $f^l_{k,p}$ for the response of the $k$'th channel at the  $p$'th location in the $l$'th convolutional layer. The
content loss $L_c$ is
\begin{equation}
L_c(I_{n}, I_{c}) = \frac{1}{2}\sum_{c} \sum_{k,p} \norm{f^c_{k,p}(I_{n}) - f^c_{k,p}(I_{c})}^2
\end{equation}
\noindent (where $c$ ranges over content layers). The {\em within-layer Gram
  matrix} for the $l$'th layer is
\begin{equation}
G_{ij}^l(I) = \sum_p \left[f_{i,p}^l(I)\right]\left[f_{j,p}^l(I)\right]^{T}
\end{equation}
\noindent Write $w_l$ for the weight applied to the $l$'th layer.  Then 
\begin{equation}
L_s^l(I_{n},I_{s})=\frac{1}{4{N^l}^2{M^l}^2}\sum_{s}w_l\sum_{i,j}\norm{G^s_{ij}(I_{n})-G^s_{ij}(I_{s})}^2
\end{equation}
\noindent where $s$ ranges over style layers. Gatys \ea use Relu1\_1, Relu2\_1, Relu3\_1, Relu4\_1, and Relu5\_1 as style layers, and layer Relu4\_2 
for the content loss, and search for $I_{n}$ using L-BFGS~\cite{liu1989limited}.  From now on, we write R51 for Relu5\_1, etc. 

\subsection{Cross-layer style loss}
We consider a style loss that takes into account between layer statistics.  The {\bf cross-layer, additive (XL)} loss
is obtained as follows.  Consider layer $l$ and $m$, both style layers, with decreasing 
spatial resolution.  
Write $\uparrow f^{m}$ for an upsampling of  $f^m$ to $H^l\times W^l \times C^m$, and consider
\begin{equation}
G_{ij}^{l,m}(I) = \sum_{p} \left[ f_{i,p}^l(I)\right]\left[\uparrow {f}_{j,p}^{m}(I)\right]^{T}
\end{equation}
as the cross-layer gram matrix, We can form a style loss
\begin{equation}
L_s(I, I_{s}) = \sum_{(l, m)\in {\cal L}} w^{l}\sum_{ij} \norm{G^{l,m}_{ij}(I)-G^{l,m}_{ij}(I_s)}^2
\end{equation}
(where ${\cal L}$ is a set of pairs of style layers).   We can substitute this loss into the original style loss, and
minimize as before.  All results here used a {\em pairwise descending} strategy, where one constrains each layer and its
successor (i.e. (R51, R41); (R41, R31); etc).  Alternatives include an {\em all distinct pairs} strategy, where one constrains all pairs of distinct
layers. Carefully controlling weights for each layer's style loss is not necessary in cross-layer gram matrix scenario.  

\section{Base Statistics for Quantitative Evaluation}\label{effcoh}

Style transfer methods should meet at least two requirements: (1) the method produces images in the desired style -- {\bf E statistics}; (2) the resulting image respects the decomposition of content image into objects 
-- {\bf C statistics}. 


{\bf Base E statistics}: We want to measure similarity of two distributions, one derived from the style image, the other from the transferred image. At each layer, e.g. R41 feature map, we first project both style image's and transferred image's summary statistics to a low-dimensional representation. Then we assume these representations are parameters of Gaussian distributions and a standard KL divergence is applied to measure the distance. The same procedure is repeated for other layers, e.g. R11,R21,R31 and R51. 

Specifically, the projection matrix at each layer is discovered as such: we first find a set of content images (we use 200 test images from BSDS500\cite{arbelaez2011contour}) $I_N=\{I_1,...,I_n\}$, and obtain their convolutional feature covariance matrices from a pretrained VGG model. Similar to the Gram matrix, a feature covariance matrix is computed by:
\begin{equation}
Cov_{ij}^l(I_n) = \sum_p \left[f_{i,p}^l(I_n) - \bar{f_{i}}^l(I_n)\right]\left[f_{j,p}^l(I_n) - \bar{f_{j}}^l(I_n)\right]^{T}
\end{equation}
where $\bar{f_{i}}^l(I_n)$ , $\bar{f_{j}}^l(I_n)$ are the $i$'th and $j$'th element of channel-wise feature mean $\bar{f}^l(I_n)$ at level $l$. Then, the average covariance matrix  $Cov_{avg}^l$ is computed by element-wise average over all images of $I_{N}$'s Covaraiance matrices at layer $l$. We apply singular value decomposition on $Cov_{avg}^l$ and keep $t$ eigenvectors corresponding the largest $t$ eigenvalues. These eigenvectors form our projection basis $P^l$ which is fixed. Given an image $I$, $I \notin I_{N}$, it's low-dimensional summary statistics at level $l$ becomes:
\begin{equation}
Mean_{proj}^l(I) = \hat{f^l(I)} P^l ; Cov_{proj}^l(I) = {P^l}^T Cov^l(I)P^l
\end{equation}

We treat $Mean_{proj}^l(I)$ and $Cov_{proj}^l(I)$ as the parameters $\mu$ and $\Sigma$ of $t$-dimensional Gaussian distribution $\mathcal{N}(\mu, \Sigma)$. $E_i$ denotes the negative $\log$  KL divergence of $i$'th layers between the transferred image  $I_{0}$ and the style image $I_{1}$, the KL distance is expressed as follow:

\begin{align}
    &D_{KL}\left(\mathcal{N}_0||\mathcal{N}_1\right)
    =\tfrac{1}{2}\left(\text{tr}\left(\Sigma^{-1}_{1}\Sigma_{0}\right)\right. \nonumber  \\ 
    &\left. +\left(\mu_{1}-\mu_{0}\right)^T\Sigma^{-1}_{1}\left(\mu_{1}-\mu_{0}\right) -t+{\text ln}\left( \tfrac{{\text det}\Sigma_1}{{\text det}\Sigma_0}\right)\right) \label{eqn:kl}
\end{align}

We reduce dimensions for two reasons:  first, we believe that image channels in feature
maps are heavily correlated; second,  a full dimension
estimate of KL divergence is likely to be dominated by variance
effects, which are particularly severe when some eigenvalues
of the covariance may be very close to zero.  For layers
R11, R21, R31, R41, R51 we use dimensions 18, 100, 128, 280, 256 respectively. 

We believe that an estimate
of the projection obtained from a sufficiently large sample
of a sufficiently rich family of images will be close to canonical (i.e. changing the
sample or the family will produce little change in projection).
This means one might reasonably estimate projection matrices using
the style images as well.    We chose to use
content images because that means the projection is not
adapted to the choice of styles (which might not be sufficiently rich).

\textbf{Base C statistics} measure the extend to which style transfer methods preserve ”objectness” in the content image. Object boundaries are a vital cue for human
perception, and we hypothesize that a transferred image that better preserves object boundaries will better reflect the content of the original image.  To measure this property, we use the off-the-shelf contour detection method by Arbelaez ~\etal \cite{arbelaez2011contour}, which estimates Pb from an image.  We use the standard metric,(the F-score, which is a harmonic
mean of precision and recall between Pb and human-drawn
contour map).
The final contour detection score is the Maximum F-score of a precision-recall curve. We compute the final contour detection scores with the transferred images' Pb and ground truth contours from the content images. The resulting contour detection scores are the base C statistics. We think this is fair because standard contour detection methods were not developed with transferred images in the scope. For source content images and human annotated ground truth contour maps we choose 200 test images from BSDS500\cite{arbelaez2011contour}.

\section{Calibrated Measures from Base Statistics}\label{calibrate}

\begin{figure*}
  \includegraphics[width=1\linewidth]{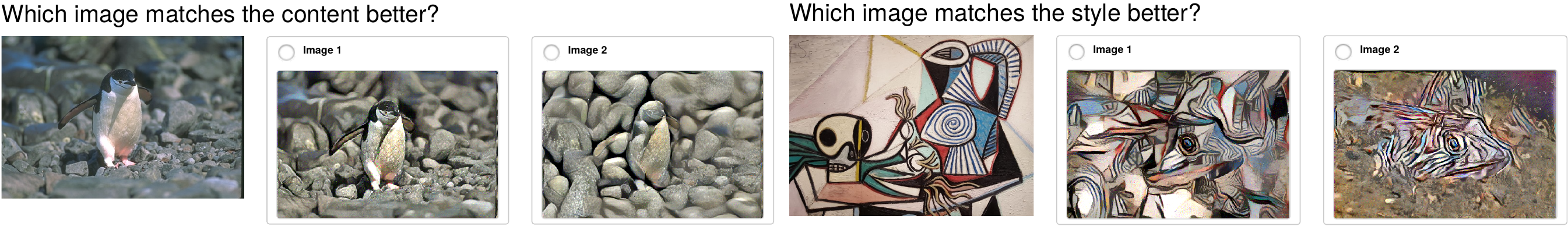}
\caption{\em 
On the {\bf left}, a typical screen from the C-test; a user must 
select which target has content most like the given content image. 
On the {\bf right}, a typical screen from the E-test; a user must 
select which target has style most like the given style image.  
  In the C-test, transferred images are selected to have reasonably
good E statistics.
\label{stus}
}
\end{figure*}

Our base EC statistics offer a quantitative measurement to style transfer methods and provide an insight in searching better style transfer methods. Yet one should calibrate with actual user preference over transferred images. Two surveys~(E-test for style and C-test for content, Fig.~\ref{stus}) can help calibrating EC statistics. 
 
In both surveys, users are presented with a pair of transferred images which only differ by style transfer methods or the same method but optimization parameters (e.g. style weights, optimization iterations), while the content and the style images are the same. In the E-test, users are asked to choose the transferred image that better captures the style. The transferred images are randomly selected from transferred results of the same style-content pair. Similarly, in the content study, users are asked to choose the image that more resemble to the content image, but the provided image pairs are chosen to have relatively high E statistics (details below). This selection is manual to ensure only seemly plausible style transferred images are used for C-test.

\subsection{Calibration with User Studies}\label{cdetail}



%

{\bf Calibration method:} {Our calibration method is mainly based on logistic regression from the base EC statistics (defined in the previous section) 
to the target human preference of user study.
Once the calibration is done, each synthesized image can have a corresponding preference score.
The difference of the scores between the two transferred images ( referred as image 1 and 2) is used to predict
 that one is preferred by the user over the other, e.g. if image 1 has score $s_1$ and image 2 has $s_2$, then the probability
 that image 1 will be preferred by a user is predicted by $e^{s_1}/(e^{s_1}+e^{s_2})$. We seek one such score for effectiveness
 (which should predict the results of the style user study) and another for coherence (which should predict the results of the content user study). }


{\bf Scores and logistic models:} Given an image pair, we have a random variable $y$ says if the image is preferred by human for a  E-test or C-test, we also have a vector of features $\vect{x}$ chosen from some combination of the base C statistic
and the 5 base E statistics. Given a pair of images ($\vect{x}_{1}$ for image 1, etc.), we can fit the logistic regression model
\begin{equation}
\frac{\log P(y_1=1|\theta, \vect{x}_{1}, \vect{x}_{2})}{\log P(y_1=0|\theta, \vect{x}_{1}, \vect{x}_{2})}=\theta^T (\vect{x}_{1}-\vect{x}_{2})
\label{eqn:logic}
\end{equation}
which yields a per-image score $s = \theta^T \vect{x}$. The choice of the admissible logistic model for user calibration is important: (a) the model should predict human preferences accurately; (b) the model should have positive weights for every base E statistics. Note that a negative weight on some feature means the model
predicts that if image 1 has a larger value of that feature than image
2, image 2 should be preferred; but our base features have the property
that an increasing value of the feature should imply a better transfer.
As a result, we believe models with negative weights cannot be trusted,
and so we require condition b.


{\bf Calibrating E statistic:}  We investigated five E-models, where the $r$'th uses $\{E_1 \ldots E_r\}$ to obtain preference scores from E-test. Table~\ref{estat} shows the
cross-validated accuracy of the models and whether they are admissible
or not. We use the admissible model with $r=3$, which has highest cross-validated accuracy; note from the standard error
statistics that accuracy differences are significant ($p<0.05$).

{\bf Calibrating C statistic:}  We investigated six C-models, where the first only uses $C$, the rest use $C$ and the $r$'th uses $\{E_1 \ldots E_{r}\}$.
Table~\ref{cstat} shows the cross-validated accuracy of the models and whether they are admissible or not. There is no
significant difference in accuracy between the two admissible models; we choose the larger model $r=1$.

\begin{figure*}
  \includegraphics[width=1\linewidth]{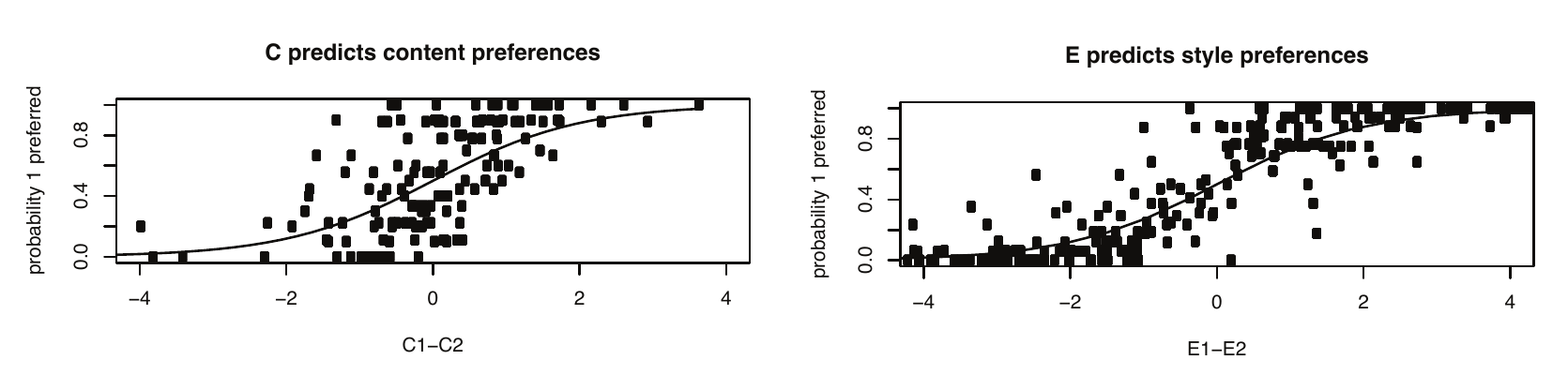}
\caption{\em 
Both E and C statistics are calibrated to user preferences in a
comparison.  On the {\bf left}, the
predicted probability of preferring image 1 to original content as a function of score $C_{1}-C_{2}$ from the selected C-model.On the {\bf right}, the predicted probability of preferring image 1 to original style as a function of score $E_{1}-E_{2}$ from the selected E-model. 
\label{calib}
}
\end{figure*}

\textbf{Visualizing calibration results}: We visualize predictions of user preference as a function of difference between scores from selected E-model and C-model in Fig.~\ref{calib}. In both plots scattered points are true user observations of style-content pairs. In the C-test each pair has 9 observations, in the E-test each pair has 16 or more observations.

\begin{table}
\begin{tabular}{|c|c|c|}
E-Model&Admissible&Cross-validated accuracy\\
1&yes&.856 (3e-3)\\
2&yes&.867 (2e-3)\\
3&yes&.873 (3e-3)\\
4&no &.871 (3e-3)\\
5&no &.873 (2e-3)\\
\end{tabular}
\caption{\em
Cross validated accuracy for our E-model predictions of human preference in the style experiment (parens give standard error of cross-validated accuracy). Model 4 and 5 are not admissible due to violating condition (b), see model description in Sec.\ref{cdetail}.  
\label{estat}
}
\end{table}

\begin{table}
\begin{tabular}{|c|c|c|}
C-Model&Admissible&Cross-validated accuracy\\
C&yes&.692 (8e-3)\\
1&yes&.694 (8e-3)\\
2&no &.710 (7e-3)\\
3&no &.756 (7e-3)\\
4&no &.759 (7e-3)\\
5&no &.767 (7e-3)
\end{tabular}
\caption{\em
Cross validated accuracy for our C-model predictions of human preference in the content experiment (parens give standard error of cross-validated accuracy).  Model 2,3,4 and 5 are not admissible due to violating condition (b), see model description in Sec.\ref{cdetail}.  
\label{cstat}
}
\end{table}

\subsection{User Study Details}
We do two rounds of user studies. The first round had 300 image pairs for E-test and 150 image pairs for C-test, each of which was generated using Gatys method\cite{gatys2016image}. In the second round, to calibrate E regardless of transfer methods, we used a mixture of 939 image pairs generated from Universal (352), XL (294) and Gatys (294) methods (see methods explanation in Sec.~\ref{ExpPro}).

\textbf{First round}: For the E-test we randomly selected two transferred images from the same style and the same content but with different optimization parameters, then paired and displayed them in random order. For the C-test we follow the same process and only used pairs where the E statistic was in the top quartile of synthesized images. For each task, users are presented with a question, an original image (style image for E-test and content image for C-test) and a transferred pair. Users are asked to choose a preferred image based on the displayed question. Overall, 16 users finished E-test, and 9 finished C-test task. From the first round we obtained 4800 clicks for E-test and 1350 clicks for C-test.

\textbf{Second round}: Only E-test was conducted at second round with the same user interface as in the first round. Different style transfer methods are applied on the same set of style-content pairs. User are provided with two transferred image using the same style-content combination but generated with different style transfer methods. 24 users (a few also participated the first round) participate the second round and contributed 2232 clicks. 

In total, from the two rounds of user study, we collected 7032 user clicks over style, and 1350 user clicks over content. Note that C-test is difficult because we selected C-test images with high E statistics. Also note that we do not evaluate on individual user preference nor on specific method, but on the correlation between general user preference and the proposed base E C statistics. Results in Tab.~\ref{estat} and ~\ref{cstat} show low standard error of mean accuracy, indicating high confidence of these experiments.    

\section{Comparing Style Transfer Methods with E and C}
With calibrated, meaningful measures of effectiveness and coherence, we can evaluate style transfer algorithms.  We
consider which algorithm is ``best'' and the effect choice of style has on performance. For analyzing the effects of
weights, choice of style,and optimization objectives etc. we use the following procedure.  We regress E and C for many style transfers produced
by the algorithm of interest, then extract information from the coefficient weights.   

\subsection{Details} \label{ExpPro}

We list style transfer methods compared in this paper:  \\
\noindent {\bf Gatys} (\cite{gatys2016image} and described above); we use the implementation by Gatys~\footnote{https://github.com/leongatys/PytorchNeuralStyleTransfer}. \\
\noindent {\bf Gatys aggressive} (\cite{gatys2016image} and described above); we use the same Gatys implementation, but with the aggressive weighting set. \\
\noindent {\bf Gatys, with histogram loss:}  as advocated by \cite{risser2017stable}, we attach a histogram loss to
Gatys method.  \\
\noindent{\bf Gatys, with layerwise style weights:}  the style weight is varied by layer;  {we multiple style losses of  layers by factors $64^{-2}$,$128^{-2}$,$256^{-2}$,$512^{-2}$,$512^{-2}$ respectively.}  \\
\noindent{\bf Gatys, with mean control:} Gatys' loss, with an {added L2 loss requiring 
that means in each transfer layer match to means in each style layer.}\\
\noindent{\bf Gatys, with covariance control:} {replacing Gatys' gram matrix by covariant matrix.}\\
\noindent{\bf Gatys, with mean and covariance control:} {replacing Gatys' style loss with  losses requiring 
that means and covariances in each layer match.\\}
\noindent {\bf Cross-layer:} We used a {\em pairwise descending} strategy with pre-trained VGG-16 model. We use R11, R21, R31,
R41, and R51 for style loss, and R42 for the content loss for style transfer.\\
\noindent {\bf Cross-layer, aggressive:} as for XL, but with the aggressive weighting set.\\
\noindent {\bf Cross-layer, multiplicative (XM):}  A natural alternative to combine style and content losses is to multiply them; we form 
$L^m(I_n) = L_c(I_n, I_c) *  L_s(I_n, I_s)$.
This provides a dynamical weighting between content loss and style loss during optimization. Although this loss function
may seem odd,  it performs extremely well in practice.\\
\noindent{\bf Cross-layer, with control of covariance (XLC)} Cross-layer loss, but replacing cross-layer gram matrices by cross-layer covariance  matrices. \\
\noindent{\bf Cross-layer, with control of mean and covariance (XLCM)}
XLC, but with an added loss requiring 
that means in each layer match.\\
\noindent {\bf Gatys, augmented Lagrangian method (GAL):}  We use the Gatys' loss, but rather than only using LBFGS to optimize,
we decouple layers to produce a constrained optimization problem and use the augmented Lagrangian method to solve this
(after the procedure in~\cite{boyd2011distributed} for decomposing MRF problems). As XM, this  works effectively as dynamical weighting
and  performs extremely well. \\
\noindent {\bf Universal Style Transfer (Universal):}(from \cite{UST}, and its Pytorch implementation \footnote{https://github.com/sunshineatnoon/PytorchWCT}.\\
\noindent {\bf Style control:} the style image is resized to content size and reported as transferred image.\\
\noindent {\bf Content control:} the content image reported as transferred image.

\vspace{-0.01cm}
 We construct a wide range of styles and contents collection, using 50 style images and the 200 content images from the BSDS500 test set. Styles are chosen by padding out the styles used in figures for previous papers with comparable images till we had 50 styles. There is not yet enough information to select a canonical style set.
We have built two dataset base on these style and content pairs. The {\em main set} is used for most experiments, and was obtained by:  
take 20 evenly spaced weight values in the range 50-2000; 
then, for each weight value, choose 15 style/content pairs uniformly and at random.  
The {\em aggressive weighting set} is used to investigate the effect of extreme weights.  This was built by taking 20 weight values sampled uniformly and 
at random between 2000-10000; then, for each weight value, 
choose 15 style/content pairs uniformly and at random.   
For each method, we then produced 300 style transfer images using each weight-style-content triplet.  
For Universal~\cite{UST}, since the maximum weight is one, 
we linearly map \textit{main set} weights to the zero-one range. 
Our samples are sufficient to produce 
clear differences in standard error bars and evaluate different
methods. 

\subsection{Results\label{results}}


\begin{figure}
\centering
  \includegraphics[width=0.9\linewidth]{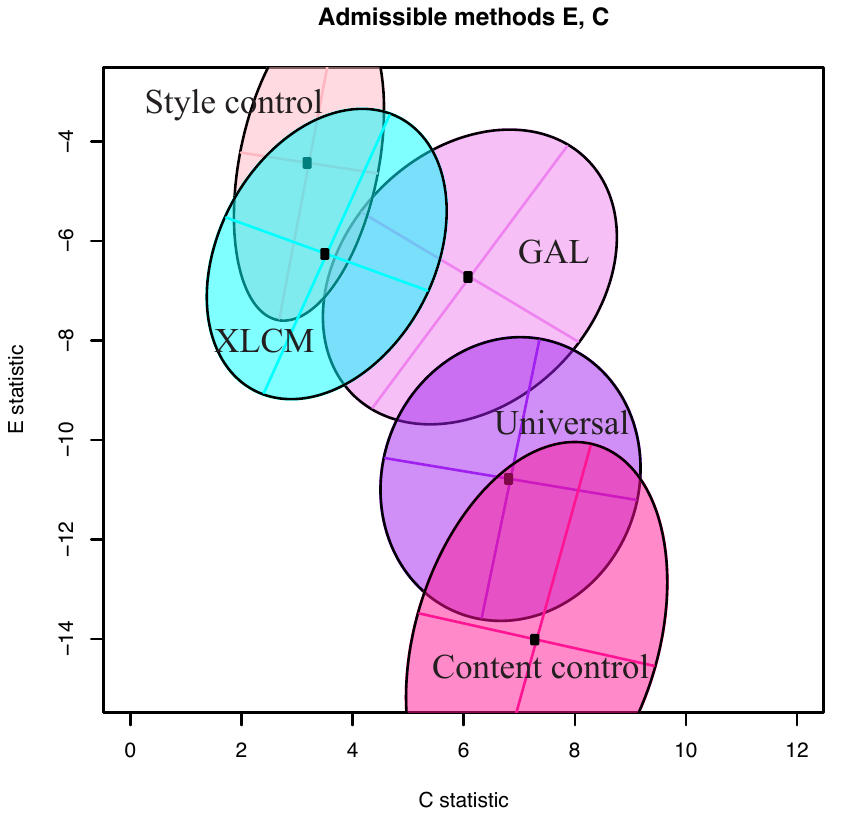}
\caption{\em
E and C statistics for admissible methods.  The plot shows mean (filled black circle) and 66\% confidence ellipse,
showing covariance of E and C values for each method.
Notice:  E and C are positively correlated, suggesting some dependence on either style (compare Fig. \protect
\ref{styledep}) or optimization difficulties; XLCM and GAL achieve better E, and universal achieves
better C;  controls are where expected (style control gets excellent E, weak C; content control weak E, excellent C).
\label{cmpadfig}
}
\end{figure}

\begin{figure}
  \includegraphics[width=0.9\linewidth]{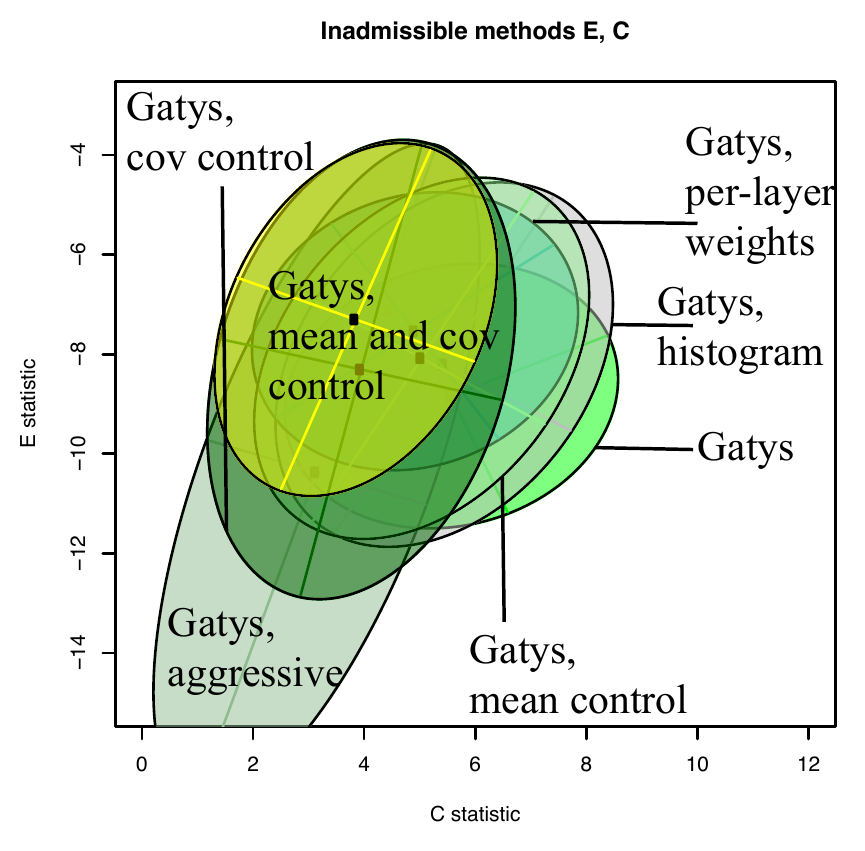}
\caption{\em
E and C statistics for inadmissible methods of the Gatys type.  The plot shows mean (filled black circle) and 66\% confidence ellipse.
Notice:  E and C are positively correlated, suggesting some dependence on either style (compare Fig. \protect
\ref{styledep}) or optimization difficulties;  the likely instability in Gatys' method is reflected by very high
variance when an aggressive weight schedule is used.
\label{cmpinadfigg}
}
\end{figure}

\begin{figure}
  \includegraphics[width=0.8\linewidth]{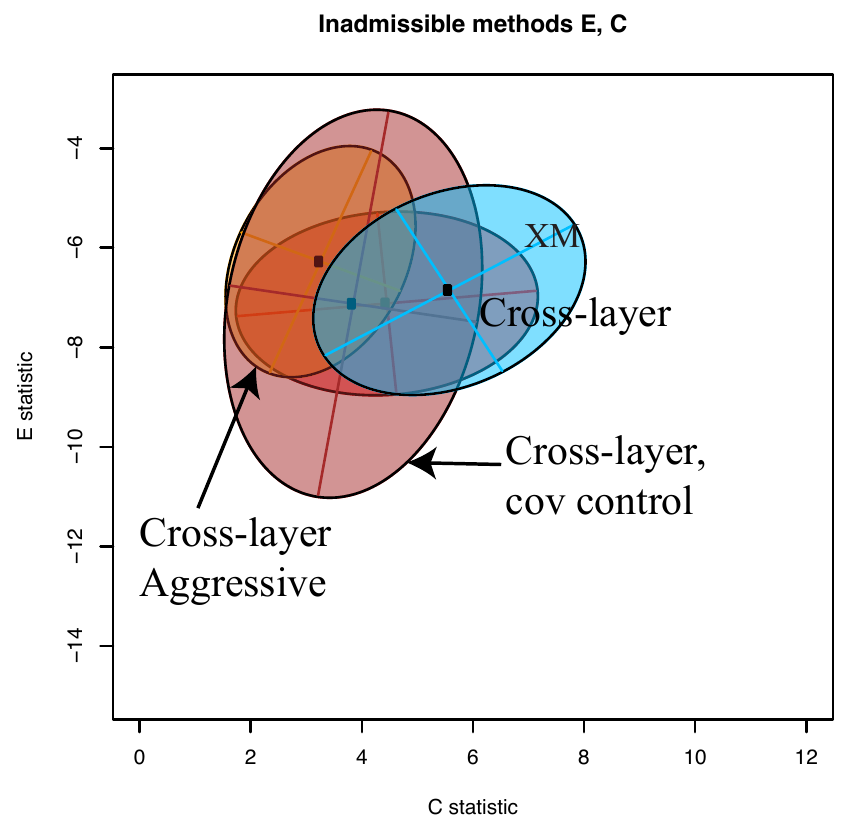}
\caption{\em
E and C statistics for inadmissible methods of the cross-layer type.  The plot shows mean (filled black circle) and 66\% confidence ellipse.
Notice:  E and C are positively correlated, suggesting some dependence on either style (compare Fig \protect
\ref{styledep}) or optimization difficulties;  the cross-layer method reacts to aggressive style weighting by producing
increased E and lower C, as one would expect.  XM performs best, and is very close to being admissible.
\label{cmpinadfigx}
}
\end{figure}

We run style transfer methods on our dataset(a tuple of {\tt style}, {\tt
  content}, and {\tt weight}), and then plot these samples with calibrated the E and C statistics for comparison.
We show the mean and covariance ellipse for E and C for various methods in Fig. ~\ref{cmpadfig}, ~\ref{cmpinadfigg} and~\ref{cmpinadfigx}. 

Generally, methods with strong C may have weak E and vice versa, which can be considered as a typical trade-off (this is a Pareto frontier).  
In spite of this trad-off phenomenon, we still can find some style methods superior than others. 
An {\bf admissible method} is a method which does not have both mean E and mean C weaker than any other methods, e.g. style control has excellent E and weak C; the content control has excellent C and weak E.  Note that this criterion is weak, because it looks at mean E and mean C, and the covariance might argue for using a method with
inadmissible means.
{Fig.~\ref{cmpadfig}} summarizes the admissible methods based on the comparison with methods shown in Fig.~\ref{cmpadfig}, ~\ref{cmpinadfigg} and~\ref{cmpinadfigx}.
Universal style transfer has excellent C, but very weak E
(i.e. the style is not much transferred, so the original image is quite coherent).   XLCM and GAL obtain only very slightly different E's, but different C's;
although each is admissible, GAL should likely be preferred as it obtains a strong C with little erosion of E.   The differences between methods quite obviously achieve statistical significance (n=300; ellipses show covariance rather than standard deviation).

{Fig.~\ref{cmpinadfigg} and~\ref{cmpinadfigx}} summarize the {\bf inadmissible methods} (for the Gatys type and the cross-layer type  respectively). Any of these methods can not beat methods  of Fig.~\ref{cmpadfig} in both mean E and mean C at same time. Note that XM is very close to being admissible.
  {Notice, in particular, that inadmissible methods tend to have large variance in C; one
might get a good C, but one might also get a bad one.}


\begin{figure*}
\centering
  \includegraphics[width=0.99\textwidth]{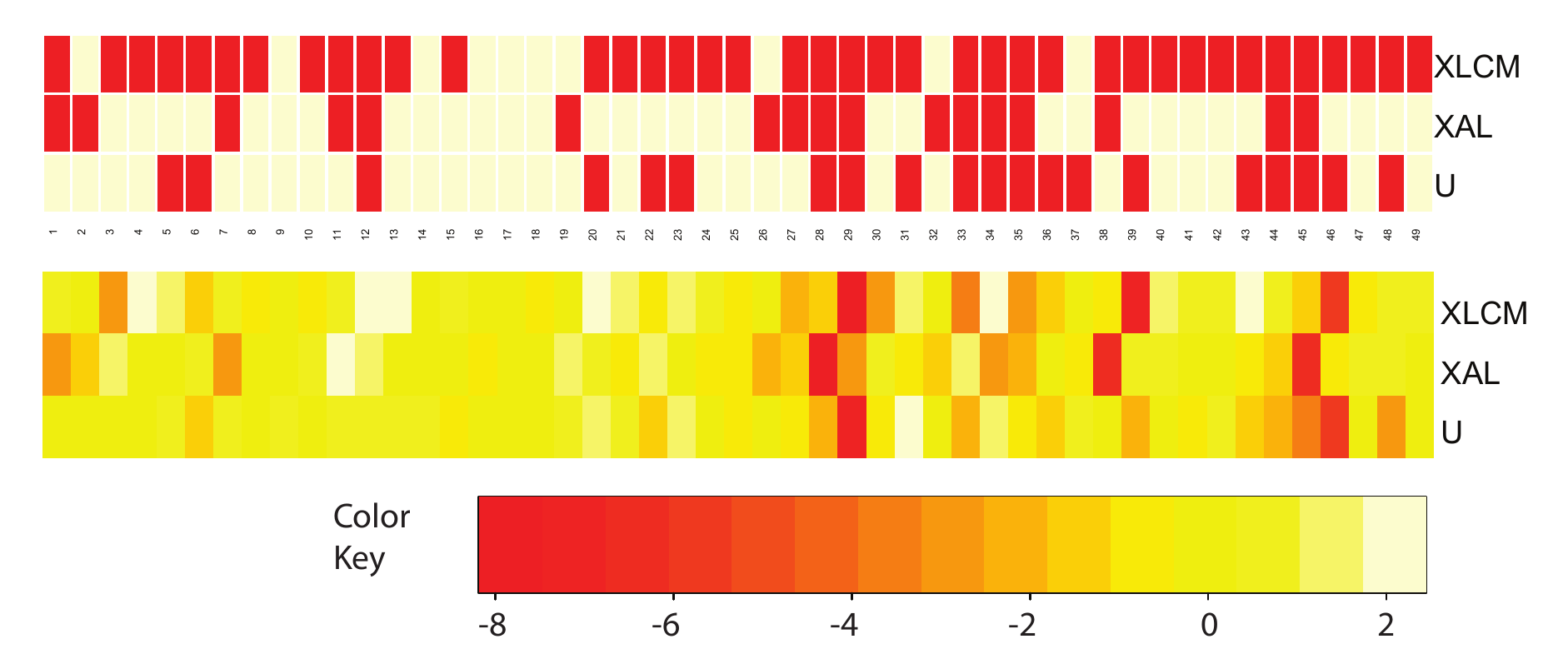}
\caption{\em
The E measure that a method produces depends very strongly on the
style; some styles transfer well, others poorly, even for admissible
methods. On
the {\bf top}, a heatmap showing the significance of the dependency of the E statistic on style, red boxes indicate
$p<0.05$ (i.e. likely not an accident).  Vertical coordinate gives the method, horizontal coordinate gives the style.
While more detailed analysis would be required to reliably identify which styles have a strong effect of the method, it
is clear that all methods are strongly affected by many styles.  On the {\bf bottom}, a heatmap showing the weight
(positive=yellow means improves E; negative=red means weakens E) for each of our 50 styles for each method.    All
methods find some styles hard and others helpful. 
\label{styledep}
}
\end{figure*}

{\bf Style and Weight:}  Style weights have surprisingly small effect
on the E statistic for admissible methods (Tab.~\ref{sweights}). Aggressive style weights lead to unstable transfer results, see \textit{Gatys, aggressive} in Fig.~\ref{cmpinadfigg} and \textit{Cross-layer, agressive} in Fig.~\ref{cmpinadfigx}.
Choice of style is very important. Fig.~\ref{styledep} shows the
result of regressing the E statistic against style identity; many
styles are strongly advantageous or disadvantageous for many methods.
There is no clearly dominant method here.  It is obvious from the figure that
any given method can be significantly advantaged by choosing the
styles for transfer carefully.  This is a trap for evaluators. 

\begin{table}
\centering
\begin{tabular}{|c|c|c|}
\hline
Admissible Method&Style Weight& Significance\\
&Effect& (P-value)\\
\hline
XLCM&-0.40 (0.23)&0.05\\
GAL&-0.34 (0.19) &0.09\\
Universal&1.54 (0.89)&$<1e-3$\\
\hline
\end{tabular}
\caption{\em We show the effect of
 style weight on E for admissible methods by multiplying the regression coefficient by the mean
  style weight (brackets show regression coefficient $\times$ standard
  deviation).  This gives the range of differences in E caused by
  style weights.  Note P-values are high for XLCM and GAL, 
 so there is little evidence weights actually matter.  
\label{sweights}
}
\vspace{-0.9cm}
\end{table}

\section{Discussion}

What causes the difference between Gatys' method and cross-layer
losses?  A {\bf symmetry analysis} \cite{risser2017stable} helps explain some aspects of our results. 
It is necessary to assume the map from layer to layer is linear.  This is not
as restrictive as it may seem; the analysis yields a local construction about any generic operating point of the
network.  In summary, we have:  The cross-layer gram matrix loss has very different symmetries to Gatys'
(within-layer) method.  In particular, the symmetry of Gatys' method can rescale
features while shifting the mean. 
For the
cross-layer loss, the symmetry cannot rescale, and cannot shift the
mean.   This implies that, if one constructs numerous style
transfers with the same style using Gatys' method, the variance of the layer features should be much greater than that
observed for the cross layer method. Furthermore, these symmetries impede optimization by making it hard to identify
progress as massive changes in the input image may lead to no change in loss.
Increasing style weights in Gatys method should result in poor
style transfers, by exaggerating the effects of the symmetry, and we observe this effect, see Gatys, aggresive in Fig.\ref{cmpinadfigg}. 

Our experimental evidence suggests the symmetries manifest themselves in
practice. Gatys-like methods displays significantly larger variance in C than cross-layer methods, and aggressive
weighting makes the situation worse. This suggests that the variance implied by the larger symmetry group is actually appearing.  In particular,
Gatys' symmetry group allows rescaling of features and shifting of their mean, which will cause the feature distribution
of the transferred image to move away from the feature distribution of
the style, causing the lower E statistic.  Histogram regularization does not appear to help significantly.

Symmetries appear to interact strongly with optimization difficulties.  GAL uses a standard optimization trick  (insert
variables and constraints to decouple terms in an unconstrained problem in the hope of making better progress with each step) and
benefits significantly.  In particular, GAL is largely immune to change in style weight.
This suggests that the main difficulty might lie with optimization procedures, rather than with losses.

\section{Conclusion}

Style transfer methods have proliferated in the absence of a quantitative evaluation method.  Our evaluation procedure attempts to provide evidents for strong style transfer methods. We calibrate out measurement to predict human preferences in style and content experiments, allowing extensive comparison of methods. Small variants on method -- for example, changes to optimization procedure -- seem to have significant effect on performance.  This is a situation where quantitative evaluation is essential.  Furthermore, our results suggest that the
choice of style strongly affects the performance of all admissible
algorithms.


{\small
\bibliographystyle{ieee}
\bibliography{egbib}
}

\section{Quick Overview}~\label{appendix_quick}

Notice that in Fig 5 all Gatys related methods except {\em Gatys with mean and covariance control} have quite low E compared to the E for cross-layer methods in Fig 6.  But {\em Gatys with mean and covariance control}  has different symmetries to Gatys (because one is controlling both mean and covariance, rather than just the Gram matrix; the symmetries are like those of the cross-layer method).   This suggests it is likely that the symmetry is at least part of the reason why some  methods outperform others.  

There are two possible reasons.  First, the symmetry results in poor solutions being easy to find.  Second, the symmetry causes optimization problems.   Both issues appear to be in play.  Figures 5 and 6 together suggest that methods have considerable variance in performance, which is consistent with poor solutions being easy to find.   But the good performance of GAL (see Fig. 4) suggests that optimization is an issue, too.

Symmetries can create problems for optimization methods, because symmetries must be associated with strong gradient curvature at least some points.
GAL uses a standard optimization trick to simplify the optimization problem; the success of this trick suggests that optimization of Gatys' loss is hard.


\subsection{GAL}
Gatys' loss is a function of feature values at each layer.  One usually assumes that the feature values taken at layer $l$ are a known function of the feature values at layer $l-1$.  Here the function is given by the appropriate convolutional layer, etc.  However, we could ``cut'' the network between layers, then introduce a constraint requiring that variables on either side of the cut be equal.  We solve this constrained problem using the augmented lagrangian method (see [4] for this strategy applied to MRFs).  

Write  $f^l_{k,p}$ for the response of the $k$'th channel at the  $p$'th location in the $l$'th convolutional layer; drop subscripts
as required,  and write $f^l=\phi^{l}(f^{l-1}_{.,.})$ for the function mapping layer to layer.   GAL cuts the layers only at R41.  We have not tried other cuts. It would be interesting to see what happened with more cuts, but the optimization problem gets big quickly.   We introduce dummy variables $V_{k, p}$, and the constraint $V=\phi^{4}(f^{3}_{.,.})$.    Write $\lambda$ for lagrange multipliers corresponding to the constraint, $I$ for the image, and $\lambda^{(i)}$ for the $i$'th estimate of those lagrange multipliers, etc.  

The augmented lagrangian is now
\begin{eqnarray*}
{\cal L}(I,V,\lambda) &= \sum_{l\neq 4}w_lL^l_{style}(I, I_{style}) \\
                      &+w_4 L_{style}^4(V,I_{style})\\
                      &+ L_{content}(V,I_{content})\\
                      &+ L_{aug}(I,  V, \lambda)
\end{eqnarray*}

where $w_l$ is the style weight of each layer, $L^l_{style}$ is the style loss for layer $l$, and $L_{content}$ is the content loss at R41, and
\begin{eqnarray*}
L_{aug}(I, V, \lambda) &= \frac{1}{KP}\sum_{k,p}\Big( \lambda_l *(V_l-\phi^4(f^{3}_{.,.}(I)))\\
                      &+ \rho(V_l-\phi^4(f^{3}_{.,.}(I)))^2 \Big)
\end{eqnarray*}

In the primal step, we first optimize the lagrangian with respect to $I$, using fixed $V$, $\lambda$ using LBFGS.
We then fix $I$, and optimize with respect to $V$ (notice this involves solving a relatively straightforward linear system). 
The dual step then re-estimates the lagrange multipliers as usual:
\[\lambda^{(i+1)}_4 =\lambda^{(i)}_4 +\rho^{(i)}(V^{(i)}_4-f^4(I^{(i)}_n)). \]
Finally, we update $\rho$ by $\rho^{(i+1)}=1.4 \rho^{(i)}$.

\begin{figure*}
\centering
  \includegraphics[width=0.5\linewidth]{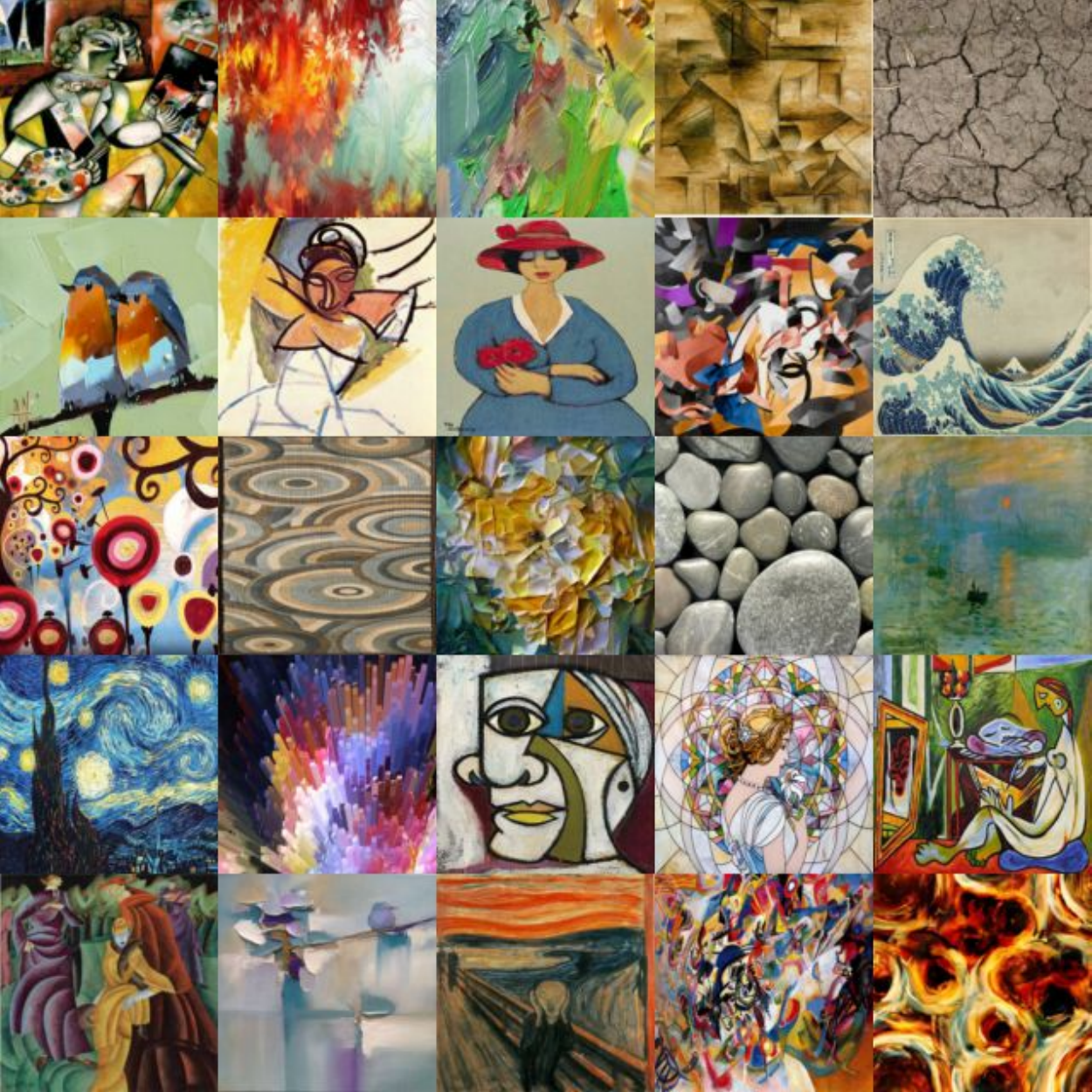}
\caption{\em The first group of 50 styles.     }
\label{50styles1}
\vspace{-3mm}
\end{figure*}

\begin{figure*}
\centering
  \includegraphics[width=0.5\linewidth]{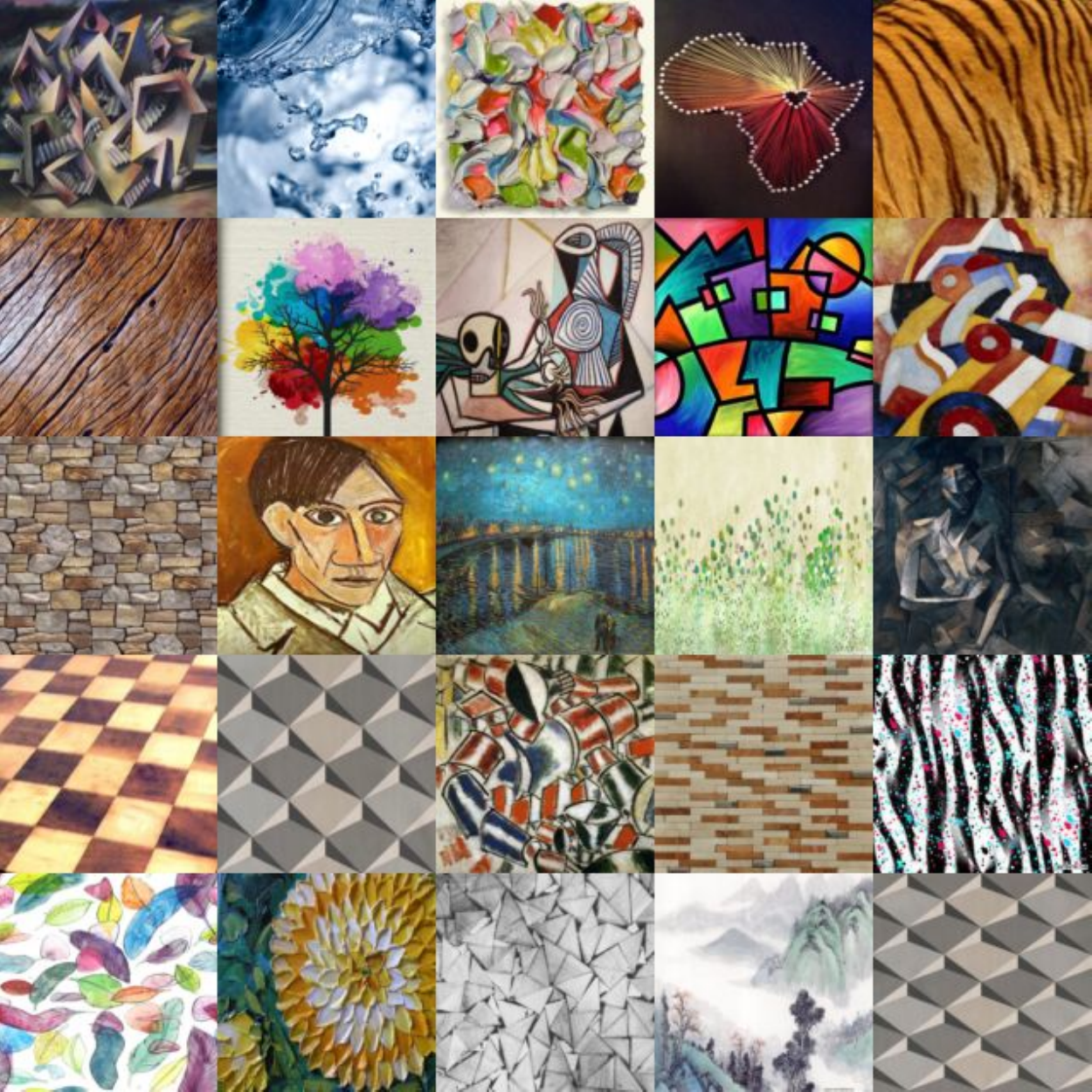}
\caption{\em The second group of 50 styles.     }
\label{50styles2}
\vspace{-3mm}
\end{figure*}

Figure \ref{50styles1} and Figure \ref{50styles2} display our 50 style images. Except the Universal style transfer, all other methods synthesize image from Gaussian noise with LBFGS optimizer. The content images and style images are resized to same width of 512 as the input for style transfers. 

\subsection{Cross-layer with control of mean and covariance (XLCM)}
We observe that feature mean difference between $I_s$ and $I_c$ is directly related to the optimization performance of style transfer, e.g. when the content image have similar feature mean as style image the transfer image has better style quality. Therefore we introduce the L2 loss between each feature channel's mean of $I_n$ and each feature channel's mean of $I_s$ to enforce the transfer image has close feature mean to style image. Here is the loss for mean control.

\[L_{mean} =  \sum_k\left(\sum_p\frac{f^l(I_n)}{P}-\sum_p\frac{f^l(I_s)}{P}\right)^2 \]%

On the other hand, the covariant control is to replace cross-layer gram matrix by corresponding cross-layer gram matrix with each feature subtracted by by its mean. Here is the new cross-layer loss with covariant control.
\[
Cov_{ij}^{l,m}(I) = \sum_{p} \left[ f_{i,p}^l(I)-\bar{f}_{i,p}^l(I)\right]\left[\uparrow {f}_{j,p}^{m}(I)-\uparrow {\bar{f}}_{j,p}^{m}(I)\right]^{T}.
\]

Here $\bar{f}_{i,p}^l(I)$ is the tensor duplicated in p dimension with the mean of $ f_{i,p}^l(I)$ over p.

\begin{table*}[t]
\begin{center}
\begin{threeparttable}
\resizebox{\textwidth}{!}{
\begin{tabular}{|p{50mm}|p{50mm}|p{50mm}|}
\hline
(low C-score, high E-score) \newline
\textbf{Cross-layer,aggressive:24.06\%}, \newline 
XLCM:20.92\%, \newline 
XLC:11.92\%,\newline 
XL:11.30\%,\newline  
GatysCM:9.21\%  
& (middle C-score, high E-score) \newline 
\textbf{XLC:14.56\%}, \newline  
Cross-layer,aggressive:13.60\%,  \newline 
XLCM:13.41\%,\newline 
XL:13.22\%,\newline 
GAL:10.15\% 
& (high C-score, high E-score) \newline 
\textbf{GAL:25.56\%}, \newline 
XM:15.04\%,\newline 
XL:10.53\%, \newline 
GatysL:8.52\%, \newline 
GatysCM:6.77\% \\
\hline
(low C-score, middle E-score) \newline
\textbf{GatysCM:15.29\%}, \newline
GatysC:12.86\%, \newline
Cross-layer, aggressive:11.65\%, \newline
GatysL:11.65\%, \newline
XLCM:8.50\% 
& (middle C-score, middle E-score) \newline 
\textbf{XM:11.69\%}, \newline
GatysM:11.49\%, \newline
GatysL:10.69\%, \newline
GatysH:10.08\%, \newline
GatysC:8.87\% 
& (high C-score, middle E-score) \newline 
\textbf{XM:15.45\%,} \newline
GatysH:14.02\%, \newline
Gatys:13.41\%, \newline
GAL:13.01\%, \newline
GatysM:11.18\% \\
\hline
(low C-score, low E-score) \newline
\textbf{Gatys aggressive:23.97\%}, \newline
GatysC:12.57\%, \newline
XLC:10.02\%, \newline
GatysCM:8.84\%, \newline
GatysM:7.47\% 
& (middle C-score, low E-score) \newline 
\textbf{Universal:12.83\%}, \newline
GatysH:10.73\%, \newline
Gatys aggressive:10.47\%, \newline
GatysM:10.21\%, \newline
Gatys:9.69\% 
& (high C-score, low E-score) \newline 
\textbf{Universal:45.28\%}, \newline
Gatys:15.75\%, \newline
GatysH:7.87\%, \newline
GatysM:6.69\%, \newline
GatysL:4.53\%\\
\hline
\end{tabular}}
\begin{tablenotes}
\item GatysH -- Gatys, with histogram loss 
\item GatysL -- Gatys, with layerwise style weights
\item GatysM -- Gatys, with mean control  
\item GatysC -- Gatys, with covariance control  
\item GatysCM -- Gatys, with mean and covariance control
\item XL -- Cross-layer
\item XM -- Cross-layer, multiplicative
\item XLC -- Cross-layer, with control of covariance
\item XLCM -- Cross-layer, with control of mean and covariance
\item GAL -- Gatys, augmented Lagrangian method 
\item Universal -- Universal Style Transfer 
\end{tablenotes}
\end{threeparttable}
\caption{Top 5 methods ranking for each quantile under regression scores coordinate generated by selected E-model and C-model. Each transferred image has five E-statistic and one C-statistic, they are used to regress user preference in E-test and C-test (Sec. 4.1 in original text). Selected E and C models regress scores (higher is better) for each transferred image. We divide the scatter into 3-by-3 quantiles, and show method distribution for each quantile.   }
\label{table:axis}
\end{center}

\end{table*}

\section{Quantization of transferred images under user study regression models }
Recall in Section 4 of original text we regress base E and C statistic to user preference. We obtain one best E-model from E-test user preference, and one best C-model from that of C-test. These two models assign E and C scores for each transferred image (Sec. 4.1 of original text). Thus, we gather a scatter plot of all transferred images, and we quantize this scatter plot into a 3-by-3 grid, each cell has roughly same number of images. From this grid we generate a visualization of EC space (Fig.1 in original text).  

This quantization shows similar trends with Figure 4-6 in the original text. Table~\ref{table:axis} shows the Top 5 methods ranking for all quantiles. In quantile of high C-score, high E-score, GAL is the top method.
XM dominates both (middle C, middle E) and (high C, middle E), and Universal dominates both (middle C, low E) and (high C, low E). Other high E quantiles are dominated by cross-layer related methods. The worst quantile(low C-score,Low E-score) has Gatys aggressive as the most popular.

This difference in symmetry groups is important.  Risser argues that
the symmetries of gram matrices in Gatys' method could lead to
unstable reconstructions; they control this effect using feature
histograms.  What causes the effect is that the symmetry rescales features while shifting the mean.  
For the cross-layer loss, the symmetry cannot rescale, and cannot shift the mean.  In turn, the instability
identified in that paper does not apply to the cross-layer gram matrix and our results could not be improved by adopting
a histogram loss.

Write $\vect{x}_{i}$, (resp $\vect{y}_i$ for the feature vector at the $i$'th location (of $N$ in total)
in the first (resp second) layer.  Write $\matx{X}^T=\left[\vect{x}_{1}, \ldots,
\vect{x}_{N}\right]$, etc.   

{\bf Symmetries of the first layer:} Now assume that the first layer has been normalized to zero mean and
unit covariance.  There is no loss of generality, because the whitening transform
can be written into the expression for the group. Write ${\cal
  G}(\matx{W})=(1/N)\matx{W}^T\matx{W}$ for the operator that forms
the within layer gram matrix. We have ${\cal G}(\matx{X})=\matx{I}$.   
Now consider an affine action on layer 1, mapping $\matx{X}_1$ to $\matx{X}_1^*=\matx{X}_1 \matx{A}+\vect{1}\vect{b}^T$; then for this to be a symmetry, we must have
$G(\matx{X}_1^*)= \matx{A}\matx{A}^T+\vect{b}\vect{b}^T=\matx{I}$.  In
turn, the symmetry group can be constructed by: choose $\vect{b}$
which does not have unit length; factor
$N(\matx{I}-\vect{b}\vect{b}^T)$ to obtain $\matx{A}(\vect{b})$ (for
example, by using a cholesky transformation); then any element of the
group is a pair $\left(\vect{b}, \matx{A}(\vect{b})\matx{U}\right)$
where $\matx{U}$ is orthonormal.  Note that factoring will fail for
$\vect{b}$ a unit vector, whence the restriction. 

{\bf The second layer:}   We will assume that the map between layers
of features is linear.  This assumption is not true in practice, but major
differences between symmetries observed under these conditions likely
result in differences when the map is linear.  We can analyze for two
cases: first, all units in the map observe only one input feature
vector (i.e. 1x1 convolutions; the {\em point sample} case); second, spatial homogeneity in the
layers.

{\bf The point sample case:} Assume that every unit in the map
observes only one input feature from the previous layer (1x1
convolutions).   We have $\matx{Y}=\matx{X}\matx{M}+\vect{1}\vect{n}^T$, because the map 
between layers is linear. Now consider the effect on the second layer.
We have ${\cal G}(\matx{Y})=\matx{M}\matx{M}^T+\vect{n}\vect{n}^T$.
Choose some symmetry group element for the first layer,
$\left(\vect{b}, \matx{A}\right)$.  The gram matrix for the second
layer becomes ${\cal G}(\matx{Y}^*)$, where
$\matx{Y}^*=(\matx{X} \matx{A}+\vect{1}\vect{b}^T)\matx{M}^T+\vect{1}\vect{n}^T.$
Recalling that $\matx{A}\matx{A}^T+\vect{b}\vect{b}^T=\matx{I}$ and
$\matx{X}^T\vect{1}=0$, we have 
\[
{\cal
  G}(\matx{Y}^*)=\matx{M}\matx{M}^T+\vect{n}\vect{n}^T+\vect{n}\vect{b}^T\matx{M}^T+\matx{M}\vect{b}\vect{n}^T
\]
so that ${\cal G}(\matx{X}_2^*)={\cal G}(\matx{X}_2)$ if
$\matx{M}\vect{b}=0$.  This is relatively easy to achieve with
$\vect{b}\neq 0$.

{\bf Spatial homogeneity:} Now assume the map between layers has
convolutions with maximum support $r \times r$.  Write $u$ for an
index that runs over the whole feature map, and $\psi(\vect{x}_u)$ for
a stacking operator that scans the convolutional support in fixed
order and stacks the resulting features. For example, given a 3x3
convolution and indexing in 2D, we might have
\[
\psi(\vect{x}_{22})=\left(\begin{array}{c}\vect{x}_{11}\\
\vect{x}_{12}\\
\ldots\\
\vect{x}_{33}
\end{array}\right)
\]

In this case, there is some $\matx{M}$, $\vect{n}$ so that 
$\vect{y}_u=\matx{M}\psi(\vect{x}_u)+\vect{n}$.  We ignore the
effects of edges to simplify notation (though this argument may go
through if edges are taken into account).  Then there is some
$\matx{M}$, $\vect{n}$ so we can write 
\[
{\cal G}(\matx{Y})=(1/N) \sum_u
\matx{M}\psi(\vect{x}_u)\psi(\vect{x}_u)^T\matx{M}^T+\vect{n}\vect{n}^T
\]
Now assume further that
layer 1 has the following (quite restrictive) spatial homogeneity
property: for pairs of feature vectors within the layer $\vect{x}_{i,j}$, $\vect{x}_{i+\delta,
  j+\delta}$ with $\mid \! \delta\!\mid \leq r$ (i.e. within a convolution window of one
another), we have $\expect{\vect{x}_{i,j}\vect{x}_{i+\delta,
    j+\delta}}=\matx{I}$.  This assumption is consistent with image
autocorrelation functions (which fall off fairly slowly), but is still
strong. Write $\phi$ for an operator
that stacks $r \times r$ copies of its argument as appropriate, so
\[\phi(\matx{I})=\left(\begin{array}{ccc}
\matx{I}&\ldots&\matx{I}\\
\ldots &\ldots \ldots \\
\matx{I}&\ldots&\matx{I}
\end{array}\right).
\]
Then
$G(\matx{Y})=\matx{M}\phi(\matx{I})\matx{M}^T+\vect{n}\vect{n}^T$.
If there is some affine action on layer 1, we have
$G(\matx{Y}^*)=\matx{M}\left(\psi(\matx{A})\phi(\matx{I})\psi(\matx{A}^T)+\psi(\vect{b})\psi(\vect{b}^T)\right)\matx{M}^T+\vect{n}\vect{n}^T$,
where we have overloaded $\psi$ in the natural way.  Now if
$\matx{M}\psi(\vect{b})=0$ and $\matx{A}\matx{A}^T+\vect{b}\vect{b}^T=\matx{I}$, ${\cal
  G}(\matx{Y}^*)={\cal G}(\matx{Y})$.

{\bf The cross-layer gram matrix:}  Symmetries of the cross-layer gram matrix are very different.  Write
${\cal G}(\matx{X}, \matx{Y})=(1/N) \matx{X}^T\matx{Y}$ for
the cross layer gram matrix.  

{\bf Cross-layer, point sample case:} Here (recalling $\matx{X}^T\vect{1}=0$)we have ${\cal G}(\matx{X},
\matx{Y})=\matx{M}^T$.    Now choose some symmetry group element for the first layer,
$\left(\matx{A}, \vect{b}\right)$.  The cross-layer gram matrix
becomes 
\begin{eqnarray*}
{\cal G}(\matx{X}^*, \matx{Y}^*)&=&(1/N)(\matx{A}
\matx{X}^T\vect{b}\vect{1}^T)\\
& &\left[(\matx{X}
  \matx{A}^T+\vect{1}\vect{b}^T)\matx{M}^T+\vect{1}\vect{n}^T\right]\\
&=&\matx{M}^T+\vect{b}\vect{n}^T
\end{eqnarray*}
(recalling that $\matx{A}\matx{A}^T+\vect{b}\vect{b}^T=\matx{I}$ and
$\matx{X}^T\vect{1}=0$).  But this means that the symmetry requires
$\vect{b}=\vect{0}$; in turn, we must have $\matx{A}\matx{A}^T=\matx{I}$.

{\bf Cross-layer, homogeneous case:} We have
\[
{\cal G}(\matx{X}, \matx{Y})=(1/N)\sum_u \vect{x}_u\left[
  \psi(\vect{x}_u)^T\matx{M}^T+\vect{n}^T\right]=\matx{M}^T.\]
Now choose some symmetry group element for the first layer,
$\left(\matx{A}, \vect{b}\right)$.  The cross-layer gram matrix
becomes 

\begin{eqnarray*}
{\cal G}(\matx{X}^*, \matx{Y}^*)&=&(1/N)\sum_u \Bigg\{ \left(\matx{A} \vect{x}_u+\vect{b}\right)\\
&+&\left[\left(\psi(\vect{x}_u)^T\psi(\matx{A}^T) + \psi(\vect{b})\right)\matx{M}^T\right.+\left.\vect{n}^T\right]\Bigg\}\\
&=&\matx{M}^T+\vect{b}\vect{n}^T
\end{eqnarray*}

(recalling the spatial homogeneity assumption, that $\matx{A}\matx{A}^T+\vect{b}\vect{b}^T=\matx{I}$ and
$\matx{X}_1^T\vect{1}=0$).  But this means that the symmetry requires
$\vect{b}=\vect{0}$; in turn, we must have
$\matx{A}\matx{A}^T=\matx{I}$.

\section{Construction of Affine Maps for Symmetry Groups}~\label{appendix_affine}

This difference in symmetry groups is important.  Risser argues that
the symmetries of gram matrices in Gatys' method could lead to
unstable reconstructions; they control this effect using feature
histograms.  What causes the effect is that the symmetry rescales features while shifting the mean.  
For the cross-layer loss, the symmetry cannot rescale, and cannot shift the mean.  In turn, the instability
identified in that paper does not apply to the cross-layer gram matrix and our results could not be improved by adopting
a histogram loss.

Write $\vect{x}_{i}$, (resp $\vect{y}_i$ for the feature vector at the $i$'th location (of $N$ in total)
in the first (resp second) layer.  Write $\matx{X}^T=\left[\vect{x}_{1}, \ldots,
\vect{x}_{N}\right]$, etc.   

{\bf Symmetries of the first layer:} Now assume that the first layer has been normalized to zero mean and
unit covariance.  There is no loss of generality, because the whitening transform
can be written into the expression for the group. Write ${\cal
  G}(\matx{W})=(1/N)\matx{W}^T\matx{W}$ for the operator that forms
the within layer gram matrix. We have ${\cal G}(\matx{X})=\matx{I}$.   
Now consider an affine action on layer 1, mapping $\matx{X}_1$ to $\matx{X}_1^*=\matx{X}_1 \matx{A}+\vect{1}\vect{b}^T$; then for this to be a symmetry, we must have
$G(\matx{X}_1^*)= \matx{A}\matx{A}^T+\vect{b}\vect{b}^T=\matx{I}$.  In
turn, the symmetry group can be constructed by: choose $\vect{b}$
which does not have unit length; factor
$N(\matx{I}-\vect{b}\vect{b}^T)$ to obtain $\matx{A}(\vect{b})$ (for
example, by using a cholesky transformation); then any element of the
group is a pair $\left(\vect{b}, \matx{A}(\vect{b})\matx{U}\right)$
where $\matx{U}$ is orthonormal.  Note that factoring will fail for
$\vect{b}$ a unit vector, whence the restriction. 

{\bf The second layer:}   We will assume that the map between layers
of features is linear.  This assumption is not true in practice, but major
differences between symmetries observed under these conditions likely
result in differences when the map is linear.  We can analyze for two
cases: first, all units in the map observe only one input feature
vector (i.e. 1x1 convolutions; the {\em point sample} case); second, spatial homogeneity in the
layers.

{\bf The point sample case:} Assume that every unit in the map
observes only one input feature from the previous layer (1x1
convolutions).   We have $\matx{Y}=\matx{X}\matx{M}+\vect{1}\vect{n}^T$, because the map 
between layers is linear. Now consider the effect on the second layer.
We have ${\cal G}(\matx{Y})=\matx{M}\matx{M}^T+\vect{n}\vect{n}^T$.
Choose some symmetry group element for the first layer,
$\left(\vect{b}, \matx{A}\right)$.  The gram matrix for the second
layer becomes ${\cal G}(\matx{Y}^*)$, where
$\matx{Y}^*=(\matx{X} \matx{A}+\vect{1}\vect{b}^T)\matx{M}^T+\vect{1}\vect{n}^T.$
Recalling that $\matx{A}\matx{A}^T+\vect{b}\vect{b}^T=\matx{I}$ and
$\matx{X}^T\vect{1}=0$, we have 
\[
{\cal
  G}(\matx{Y}^*)=\matx{M}\matx{M}^T+\vect{n}\vect{n}^T+\vect{n}\vect{b}^T\matx{M}^T+\matx{M}\vect{b}\vect{n}^T
\]
so that ${\cal G}(\matx{X}_2^*)={\cal G}(\matx{X}_2)$ if
$\matx{M}\vect{b}=0$.  This is relatively easy to achieve with
$\vect{b}\neq 0$.

{\bf Spatial homogeneity:} Now assume the map between layers has
convolutions with maximum support $r \times r$.  Write $u$ for an
index that runs over the whole feature map, and $\psi(\vect{x}_u)$ for
a stacking operator that scans the convolutional support in fixed
order and stacks the resulting features. For example, given a 3x3
convolution and indexing in 2D, we might have
\[
\psi(\vect{x}_{22})=\left(\begin{array}{c}\vect{x}_{11}\\
\vect{x}_{12}\\
\ldots\\
\vect{x}_{33}
\end{array}\right)
\]

In this case, there is some $\matx{M}$, $\vect{n}$ so that 
$\vect{y}_u=\matx{M}\psi(\vect{x}_u)+\vect{n}$.  We ignore the
effects of edges to simplify notation (though this argument may go
through if edges are taken into account).  Then there is some
$\matx{M}$, $\vect{n}$ so we can write 
\[
{\cal G}(\matx{Y})=(1/N) \sum_u
\matx{M}\psi(\vect{x}_u)\psi(\vect{x}_u)^T\matx{M}^T+\vect{n}\vect{n}^T
\]
Now assume further that
layer 1 has the following (quite restrictive) spatial homogeneity
property: for pairs of feature vectors within the layer $\vect{x}_{i,j}$, $\vect{x}_{i+\delta,
  j+\delta}$ with $\mid \! \delta\!\mid \leq r$ (ie within a convolution window of one
another), we have $\expect{\vect{x}_{i,j}\vect{x}_{i+\delta,
    j+\delta}}=\matx{I}$.  This assumption is consistent with image
autocorrelation functions (which fall off fairly slowly), but is still
strong. Write $\phi$ for an operator
that stacks $r \times r$ copies of its argument as appropriate, so
\[\phi(\matx{I})=\left(\begin{array}{ccc}
\matx{I}&\ldots&\matx{I}\\
\ldots &\ldots \ldots \\
\matx{I}&\ldots&\matx{I}
\end{array}\right).
\]
Then
$G(\matx{Y})=\matx{M}\phi(\matx{I})\matx{M}^T+\vect{n}\vect{n}^T$.
If there is some affine action on layer 1, we have
$G(\matx{Y}^*)=\matx{M}\left(\psi(\matx{A})\phi(\matx{I})\psi(\matx{A}^T)+\psi(\vect{b})\psi(\vect{b}^T)\right)\matx{M}^T+\vect{n}\vect{n}^T$,
where we have overloaded $\psi$ in the natural way.  Now if
$\matx{M}\psi(\vect{b})=0$ and $\matx{A}\matx{A}^T+\vect{b}\vect{b}^T=\matx{I}$, ${\cal
  G}(\matx{Y}^*)={\cal G}(\matx{Y})$.

{\bf The cross-layer gram matrix:}  Symmetries of the cross-layer gram matrix are very different.  Write
${\cal G}(\matx{X}, \matx{Y})=(1/N) \matx{X}^T\matx{Y}$ for
the cross layer gram matrix.  

{\bf Cross-layer, point sample case:} Here (recalling $\matx{X}^T\vect{1}=0$)we have ${\cal G}(\matx{X},
\matx{Y})=\matx{M}^T$.    Now choose some symmetry group element for the first layer,
$\left(\matx{A}, \vect{b}\right)$.  The cross-layer gram matrix
becomes 
\begin{eqnarray*}
{\cal G}(\matx{X}^*, \matx{Y}^*)&=&(1/N) (\matx{A}
\matx{X}^T+\vect{b}\vect{1}^T) \\
& &
\left[(\matx{X}
  \matx{A}^T+\vect{1}\vect{b}^T)\matx{M}^T+\vect{1}\vect{n}^T\right]\\
&=&\matx{M}^T+\vect{b}\vect{n}^T
\end{eqnarray*}
(recalling that $\matx{A}\matx{A}^T+\vect{b}\vect{b}^T=\matx{I}$ and
$\matx{X}^T\vect{1}=0$).  But this means that the symmetry requires
$\vect{b}=\vect{0}$; in turn, we must have $\matx{A}\matx{A}^T=\matx{I}$.

{\bf Cross-layer, homogeneous case:} We have
\[
{\cal G}(\matx{X}, \matx{Y})=(1/N)\sum_u \vect{x}_u\left[
  \psi(\vect{x}_u)^T\matx{M}^T+\vect{n}^T\right]=\matx{M}^T.\]
Now choose some symmetry group element for the first layer,
$\left(\matx{A}, \vect{b}\right)$.  The cross-layer gram matrix
becomes 

\begin{eqnarray*}
{\cal G}(\matx{X}^*, \matx{Y}^*)&=&(1/N)\sum_u \Bigg\{ \left(\matx{A} \vect{x}_u+\vect{b}\right)\\
&+&\left[\left(\psi(\vect{x}_u)^T\psi(\matx{A}^T) + \psi(\vect{b})\right)\matx{M}^T\right.+\left.\vect{n}^T\right]\Bigg\}\\
&=&\matx{M}^T+\vect{b}\vect{n}^T
\end{eqnarray*}

(recalling the spatial homogeneity assumption, that $\matx{A}\matx{A}^T+\vect{b}\vect{b}^T=\matx{I}$ and
$\matx{X}_1^T\vect{1}=0$).  But this means that the symmetry requires
$\vect{b}=\vect{0}$; in turn, we must have
$\matx{A}\matx{A}^T=\matx{I}$.

\end{document}